\documentclass{main}
\usepackage{bm}
\usepackage{array}
\usepackage{subfigure}
\usepackage{multirow}
\usepackage[ruled,linesnumbered]{algorithm2e}
\usepackage{stfloats}
\usepackage{amsmath}

\title{\vspace{-4cm}Rethinking Classifier and Adversarial Attack}

\author{
{Youhuan Yang $^{1,2}$,}
{Lei Sun \xff$^{2}$,}
{Leyu Dai $^{2}$,}
{Song Guo $^{2}$,}
{Xiuqing Mao $^{2}$,}
{Xiaoqin Wang $^{2}$,}
{Bayi Xu $^{1,2}$}
}

\address{{1\quad School of Cyber Science and Engineering, Zhengzhou University, Zhengzhou 450000, China}\\
{2\quad Information Engineering University, Zhengzhou 450001, China}\vspace{3mm}}

\begin{document}
\maketitle
\setcounter{page}{1}
\setlength{\baselineskip}{14pt}

\begin{abstract}
    Various defense models have been proposed to resist adversarial attack 
    algorithms, but existing adversarial robustness evaluation methods 
    always overestimate the adversarial robustness of these models (i.e., 
    not approaching the lower bound of robustness). To solve this problem, 
    this paper uses the proposed decouple space method to divide the classifier 
    into two parts: non-linear and linear. Then, this paper defines the 
    representation vector of the original example (and its space, i.e., 
    the representation space) and uses the iterative optimization of Absolute 
    Classification Boundaries Initialization (ACBI) to obtain a better attack 
    starting point. Particularly, this paper applies ACBI to nearly 50 widely-used 
    defense models (including 8 architectures). Experimental results show that 
    ACBI achieves lower robust accuracy in all cases.
\end{abstract}

\section{Introduction}
    
    \noindent With the rapid development of artificial intelligence, various deep 
    learning/machine learning methods \cite{1,2,3,4,5} are widely used in the fields 
    of face recognition, automatic driving, malicious traffic detection, 
    medical image processing, object detection \cite{6} etc. Although deep learning methods have 
    shown powerful performance, research \cite{7} shows that deep neural networks 
    are extremely vulnerable to adversarial attacks. Adversarial attacks 
    make the model output unreasonable predictions by adding slight 
    perturbations that are difficult to distinguish by human eyes to 
    the input of the neural network. Meanwhile, various defense methods have 
    been proposed to resist adversarial attack algorithms. However, these 
    methods perform suboptimally in face of more advanced attack algorithms. 
    Therefore, model robustness evaluation methods are necessary to evaluate 
    the adversarial robustness of various defense algorithms.

    The adversarial robustness evaluation method of the trained model specifies 
    the test data set and a certain number of iterations to find its worst 
    performance. Under such task requirements, white-box attack algorithms 
    (assuming that the attacker knows all the information of the model) 
    are widely used to evaluate the adversarial robustness of defense 
    models because of their strong attack performance, e.g., Fast Gradient 
    Sign Method (FGSM)\cite{8}, Basic Iterative Method (BIM) \cite{9}, Projected 
    Gradient Descent (PGD) \cite{10}, Carlini \& Wagner (CW) \cite{11}, Fast Adaptive 
    Boundary Attack (FAB) \cite{12}, etc. Adversarial attacks could be 
    expressed as generating an adversarial perturbation $\delta$ on a 
    given example pair $\left(x,y\right)$ such that the output of the 
    model $f$ is $f\left(x+\delta\right)\neq y$. The adversarial disturbance $\delta$ 
    is generally generated by multi-step iteration, and to enhance the 
    strength of adversarial attacks, many algorithms do not generate 
    adversarial examples from the original example $x$ ($x\in\left[0,1\right]^{D}$) 
    as the starting point. Instead, they use the initialization strategy 
    to generate an initial $\delta_0$ ($\delta_0\in[-\epsilon ,+\epsilon]^{D}$) and then 
    perform an iterative attack (such as PGD, etc.) on $x_{st}=x+\delta_0$, 
    where $\epsilon$ is an important hyperparameter that restricts the 
    adversarial disturbance to the naked eye, i.e., $\epsilon$-ball 
    constraint (sometimes called box-constraint). Although 
    this random strategy enables the attack algorithm to evaluate 
    the adversarial robustness of the defense model from multiple 
    directions, it is suboptimal because it ignores the relationship 
    between the attack starting point $x_0$ and model classification 
    boundaries that need adversarial robustness evaluation. Through 
    experiments, it is found that the closer the starting point 
    of the attack algorithm is to the nearest classification boundary 
    of the trained model, the easier it is to generate adversarial 
    examples.
    
    The contributions of this paper are summarized as follows:
    \begin{itemize}

        \item This paper uses proposed decouple space to divide the 
        classifier $f$ into two sub-models $f_h$ and $f_t$, where $f_t$ stands 
        for the last layer at the tail of the neural network (fully connected 
        layer, and the number of output neurons is the number of categories), 
        and $f_h$ stands for all other layers in the head. This paper 
        defines the space between $f_h$ and $f_t$ as a representation space 
        (the vector under this space is called the representation vector, 
        which represents an example point in the original space). Using this 
        partitioning method, this paper divides the classifier prediction 
        problem into two problems (a non-linear problem and a linear problem) 
        and gives an expression for the absolute classification boundaries of 
        any classifier.
        \\
        \item This paper proposes a new adversarial perturbation initialization 
        strategy called ACBI based on absolute classification boundaries. ACBI optimizes the 
        distance between the representation vector and the 
        nearest absolute classification boundary by the gradient descent method 
        to obtain a better attack starting point, thereby improving the reliability 
        (approaching the lower bound of robustness) and efficiency (needs fewer 
        iterations to achieve a successful attack) of the existing adversarial 
        robustness evaluation methods.
        \\
        \item This paper evaluates adversarial robustness by using ACBI-based 
        adversarial attack algorithms on more than 50 defense methods. 
        Experiments show that ACBI makes different defense models achieve 
        lower accuracy under different attack algorithms.
    
    \end{itemize}

\section{Preliminaries}

    \noindent Many adversarial attack algorithms, e.g., FGSM, BIM, and other 
    algorithms, attack from the original example point. By contrast, PGD, 
    FAB, etc. attack from a random starting point near the original example 
    point each time they restart the attack. This paper selects PGD and FAB 
    attack algorithms for subsequent discussions and experiments because the 
    attack processes and loss functions used by these two algorithms are quite 
    different. Also, other adversarial attack algorithms are more or less similar 
    to PGD and FAB, e.g., FGSM, BIM, CW, the improved versions of PGD, etc. are 
    similar to PGD, and DeepFool \cite{13} and AutoAttack \cite{14} are similar to FAB.

    The adversarial attack process of PGD is as follows:
        \begin{equation}
            x_{adv}^{t+1}={{Proj}_{(x,\epsilon)}(x}_{adv}^t+\alpha{sign(\mathrm{\nabla}}_{x_{adv}^t}\mathcal{L}(f(x_{adv}^t),y)))
        \end{equation}
    where,
        \begin{equation*}
            x_{adv}^0=x_{orig}+\delta_0
        \end{equation*}
        \begin{equation*}
            x_{orig},x_{adv}^0\in\left[0,1\right]^D
        \end{equation*}
        \begin{equation*}
            \delta_0\in[-\epsilon,+\epsilon]^D
        \end{equation*}
    
    In Eq. (1), $sign$ is the sign function; $t$ is the $t$-th attack iteration 
    number; $x_{orig}$ denotes the original image; $\delta_0$ denotes the randomly 
    initialized noise; ${Proj}_{(x,\epsilon)}$ clips the input to the $\epsilon$-ball 
    of x; $\alpha$ is the step size of a single attack; $\mathcal{L}$ denotes 
    the loss function.

    The process of generating adversarial examples using FAB is as follows:
        \begin{equation}
            x_{adv}^{t+1}={Proj}_C((1-\beta)(x_{adv}^t+\eta\delta_{adv}^t)+\beta(x_{orig}+\eta\delta_{orig}^t))
        \end{equation}
    where,
        \begin{equation*}
            x_{adv}^0\gets x_{orig}\ or\ x_{adv}^0\gets random\ sampled\ s.t.\ {||x_{adv}^0-x_{orig}||}_p=\mu
        \end{equation*}
        \begin{equation*}
            \delta={Proj}_p(x,\pi_s,C)-x
        \end{equation*}
        \begin{equation*}
            \beta=min{\frac{{||\delta_{adv}^t||}_p}{{||\delta_{adv}^t||}_p+{||\delta_{orig}^t||}_p},{\ \beta}_{max}}\in[0,\ 1]
        \end{equation*}
    
    In Eq.(2), $\beta$ is a balance parameter using historical information; $\eta$ 
    is a hyperparameter that controls the size of the perturbations; $\mu$ 
    is a parameter that limits the initial size of the starting point; ${||\bullet||}_p$ 
    denotes $L-p$ norm; ${Proj}_C$ is the projection onto the box (or box constraints), 
    and it can be realized by clipping; operation ${Proj}_p(x,\pi_s,C)$ will 
    projects x onto $\pi_s$ and then clips value to the box-constraint; $\pi_s$ 
    represents the approximate hyperplane at the example point $x_{adv}^t$. For 
    details, please refer to \cite{12}.

    From the calculation process of PGD and FAB, it can be found that $x_{adv}^0$ 
    is randomly selected from the vicinity of $x_{orig}$ as the starting point 
    to attack (of course, $x_{adv}^0$ needs to meet the box-constraint). 
    Obviously, these starting points are suboptimal. To handle this 
    problem, this paper proposes an ACBI strategy, and the implementation 
    details will be presented in the next section.

\section{ACBI: Absolute Classification Boundaries Initialization}

\subsection{Revisit the classifier}

    \noindent \textbf{Traditional perspective}: generally, a classifier is composed 
    of multiple network layers (convolution, pooling, fully connected, 
    non-linear activation functions, etc.). The non-linear activation 
    function (such as ReLu, Sigmoid, etc.) is the key to the non-linearity 
    of one classifier.
        \begin{equation}
            z=f(x)=f_{...}(f_2(f_1(x)))
        \end{equation}
    where $f_i$ is a single network layer in the classifier. Many researchers 
    treat the classifier as $f$ (obviously $f$ is non-linear) and discuss various 
    problems of the classifier as a whole. In this case, many linear solutions 
    cannot be directly applied to $f$. Therefore, studies \cite{8,12,15} assume 
    that $f$ can be approximately regarded as a linear expression. However, these 
    studies are all suboptimal.

    \textbf{New perspective}: first of all, this paper ignores the softmax (or 
    sigmoid) function during classifier inference. This is because these 
    functions are often used to normalize the output logits of the model 
    as the predicted probability, which is strictly monotonically increasing. 
    Also, the presence or absence of these functions will not affect the Absolute 
    Classification Boundaries  of the classifier (this will be explained in the 
    next section). Therefore, this paper does not discuss the case where the final 
    output of the classifier passes through activation functions such as softmax 
    without any special instructions.

    This paper splits the classifier into two parts, $f_h$ and $f_t$, where $f_t$ 
    represents the last layer of the classifier (a fully connected layer, 
    the output dimension is $K$, i.e., $K$ categories for a classification task), 
    and $f_h$ represents all other network layers of the classifier (except 
    the last layer). This paper refers to this as decouple space. Given 
    an example pair ($x$,$y$) ($x$ is the model input and $y$ is the one-hot 
    encoded example label), then the output of the classifier 
    $z$ ($z\in\mathbb{R}^K$, $z_i$ is the score given by the classifier 
    on the $i$-th class) can be expressed as:
        \begin{equation}
            z=f\left(x\right)=f_t\left(v\right)=wv+b
        \end{equation}
    where,
        \begin{equation}
            v=f_h\left(x\right)
        \end{equation}
    
    This paper regards $v$ as a representation vector of example point $x$ in 
    $N$-dimensional space ($v\in\mathbb{R}^N$, $w\in\mathbb{R}^{(K,N)}$, 
    $b\in\mathbb{R}^K$). It is worth noting that multiple example points in 
    the $N$-dimensional space may correspond to a representation vector. It 
    is difficult to reverse map from $v$ back to $x$ due to the existence of 
    irreversible operations such as ReLu \cite{16}, MaxPooling  etc.
        \begin{equation}
            x\overset{f_{h}}{\rightarrow}v\overset{f_{t}}{\rightarrow}z
        \end{equation}
    
    Thus, this paper divides the non-linear $f$ into two parts following 
    Eq. (6) while maintaining the non-linear $f_h$ and linear $f_t$. All linear 
    analysis methods and solutions can be applied when only the linear 
    classifier $f_t$ is considered and $f_h$ is ignored.

\subsection{Decision boundary}

    \noindent \textbf{Fuzzy Boundary}: in the traditional view, the notion of boundaries 
    is ambiguous because the classifier is highly non-linear. Many researchers 
    choose to use the input $x$ and output $z$ to approximately fit some curves 
    \cite{17,18,19} and regard these curves as the classification boundaries of the 
    classifier. However, this approach has two disadvantages: 1) the quality 
    of the curve fitting and the degree of deviation from the true classification 
    boundary depend on the prediction accuracy of the classifier and its 
    generalization. 2) the exact expression of the classification boundaries 
    cannot be given, so these boundaries cannot be used to solve some problems. 
    In view of these disadvantages, some researchers adopt a suboptimal solution, 
    treat the non-linear classifier as a linear map approximately \cite{8,12}, and 
    apply some linear analysis to solve problems. Of course, these methods work 
    well when the size of the classifier is small (e.g. MLP or Small CNN), but 
    the approximate linearity is poor for the current mainstream large or huge 
    classifiers (e.g. ResNet-18 \cite{2} or WRN34-10 \cite{5}).

    \textbf{Absolute Boundary}: this paper proposes Absolute Boundaries  based on the decouple 
    space, which only uses $f_t$ to calculate the expression of absolute boundaries 
    (different from fuzzy boundaries, the independent variable of absolute boundaries 
    expression is the representation vector $v$ rather than the original example $x$). 
    The calculation of these expressions is introduced as follows.

    Assuming that example $x$ belongs to the $i$-th class, the requirements for making 
    the classifier's prediction result correct are as follows:
        \begin{equation}
            i=argmax\ z
        \end{equation}
    and
        \begin{equation*}
            \forall j\in\left[0,K\right),j\neq i
        \end{equation*}

        \begin{align}
            F_{(i,j)}(v) &=z_i-z_j>0 \nonumber \\
            &=w_iv+b_i-(w_jv+b_j)>0 \nonumber \\
            &=(w_i-w_j)v+(b_i-b_j)>0 \nonumber \\
            &=w_{(i,j)}v+b_{(i,j)}>0
        \end{align}
    
    It can be seen that non-linear normalization functions such as softmax will 
    not affect the results of Eq. (7) and Eq. (8), so this paper ignores these 
    non-linear factors. In Eq. (8), $w_{(i,j)}$ is the difference vector of the 
    $i$-th and $j$-th dimension vectors of the weight parameter $w$. The above formula 
    indicates that a set of inequality equations actually determines which 
    category the classifier predicts. There are $K-1$ inequality equations in 
    total, and the prediction score $z_i$ of the $i$-th dimension needs to be 
    larger than the prediction scores $z_j$ of all other categories. The same 
    is true for predicting other categories.
    
    For the convenience of discussion, this paper ignores the case where the 
    values of the two dimensions in the prediction $z$ are the same and both 
    are the largest (corresponding to $F_{(i,j)}(v)=0$ in Eq. (8)), this 
    situation is almost impossible and is not helpful for our analysis. 
    Thus, $F_{(i,j)}(v)>0$ is a necessary and insufficient condition for 
    the classifier to predict the $i$-th class (of course, $F_{(i,j)}(v)<0$ 
    is a necessary and insufficient condition for the classifier to 
    predict the $j$-th class). This paper defines $F_{(i,j)}(v)=0$ as the 
    classification decision equation of the classifier for the $i$-th 
    class and the $j$-th class. According to the above discussion, this 
    paper gives the necessary and sufficient conditions for the classifier 
    to predict the $i$-th class as (i.e., the equivalent expression of Eq. (7)):
        \begin{equation}
            F_i(v)>0\ \gets\bigcap_{i,j\in[0,K),i\neq j}{F_{(i,j)}(v)>0}
        \end{equation}

    The set of equations composed of classification decision equations 
    between all categories($i\neq j$) is:
        \begin{equation}
            \begin{split}
                F&=\{ F_{\left(0,1\right)}(v)=0,\ldots,F_{\left(i,j\right)}(v)=0,\ldots,F_{\left(K-2,K-1\right)}(v)=0 \} \\
                &=\{ {w}_{\left(0,1\right)}v+b_{\left(0,1\right)}=0,...,\\
                &\ \ \ \ \ w_{\left(i,j\right)}v+b_{\left(i,j\right)}=0,...,\\
                &\ \ \ \ \ w_{\left(K-2,K-1\right)}v+b_{\left(K-2,K-1\right)}=0 \}\\
                &=\{\mathcal{W}v+\mathcal{B}=0\}
            \end{split}
        \end{equation}
    
    It is worth noting that $F_{\left(i,j\right)}(v)=0\Leftrightarrow F_{\left(j,i\right)}(v)=0$, 
    $F_{\left(i,j\right)}(v)>0\Leftrightarrow F_{\left(j,i\right)}(v)<0$. Therefore, 
    $F$ contains a total of $C_K^2$ (i.e., $K(K-1)/2$) classification decision equations 
    (a total of $K-1$ ones related to the $i$-th class). So, for any $v\in\mathbb{R}^N$, 
    $F$ divides $\mathbb{R}^N$ into $K$ subspaces (corresponding to $K$ categories), and 
    there is no intersection between any two spaces, that is:
        \begin{equation*}
            \forall v\in\mathbb{R}^N
        \end{equation*}
        \begin{equation*}
            \mathbb{R}^N=\bigcup_{k=0}^{K-1}S_k(S_i\cap S_j=\emptyset \ i,j\in[0,K]\ i\neq j)
        \end{equation*}
        \begin{equation}
            s.t.\ v\in S_k(or\ F_k(v)>0)
        \end{equation}
    
    The above definitions of the classification decision equations and the 
    division of the representation vector spaces ($v\in\mathbb{R}^N$) are curical. 
    All problems in this space are linear, and all the following discussions 
    are conducted in this representation vector space.

\subsection{How adversarial attacks work}
    
    \noindent Although various deep learning methods show strong performance, 
    Szegedy et al. \cite{7} found that adding a slight perturbation that is 
    difficult to distinguish with the naked eye to the input image can 
    fool the classifier and make it obtain wrong prediction results. For 
    any example $x$ (assuming it belongs to class $y$), the adversarial attack 
    can be expressed as follows:
        \begin{gather}
            y\neq argmax\ {f(x+\delta)}\\
            s.t.\ x+\delta\in[0,1]^{D},\ ||\delta ||_{p}\leq \epsilon \notag
        \end{gather}
    where $\delta$ denotes adversarial perturbation, and ${||\bullet||}_p$ denotes 
    the $L-p$ distance used to limit the size of the adversarial perturbation. 
    This paper uses the commonly used $L-\infty$ attack method 
    (${||\delta||}_\infty\le\epsilon$) without special instructions.

    Under the operation of the decouple space, adversarial attacks can be 
    expressed in the representation space as follows:
        \begin{gather}
            x+\delta\overset{f_h}{\rightarrow}v+\Delta v\\
            F_k(v+\Delta v)>0(or\ v+\Delta v\in S_k) and k\neq y,\ k\in[0,\ K) \nonumber
        \end{gather}
    
    In this way, adversarial attacks actually add a slight perturbation $\delta$ 
    to the original input $x$ so that there is an increment $\Delta v$ in the 
    representation space, and $\Delta v$ makes $v$ located in the non-label domain 
    (i.e., $v+\Delta v \notin S_y$). Meanwhile, when $k$ in Eq. (13) is designated 
    manually, it is a target attack; otherwise, it is a non-target attack.

    Fig. 1 shows the attack scenario of the adversarial attack in the representation 
    space. The attacked model is a small convolutional neural network (two 
    convolutional layers followed by BatchNormalization \cite{20}, ReLu and 
    MaxPooling  layers, and the size of the output channels are 16 and 32, 
    respectively; then, followed by two fully connected layers with output 
    units of $N$ and $K$ respectively). The model is trained  with the MNIST data. 
    In particular, the experiment shown in Fig. 1 only selected four categories 
    of pictures (number 0, 1, 2, and 3) in partial MNIST, and the representation 
    space is reduced to 2 dimensions, mainly to make the representation space 
    displayable ($N=2$) and reduce the clutter of elements in the space ($K=4$).
        \begin{figure*}[!ht]
            \centering
            \includegraphics[width=1.0\textwidth]{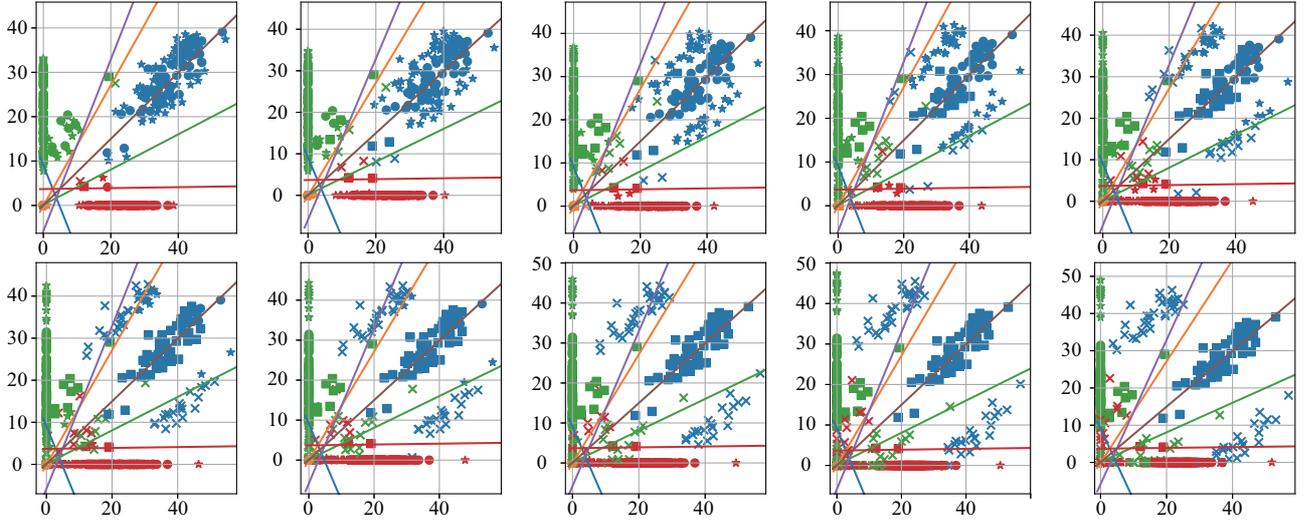}
            \vspace{-1em}
            \caption{Effects of BIM on representation vectors on partial MNIST}
            \label{Fig01}
        \end{figure*}

    Fig. 1 shows the attack effect of BIM-10 at each step on partial MNIST, 
    where each line represents the absolute classification decision boundary 
    between any two categories obtained by Eq. (8). The color of all points 
    represents the true category to which they belong, and the shape of the 
    points represents a different type of example. $\square$ and $\bullet$ represent the 
    original example points, and $\times$ and $\bigstar$ represent the adversarial examples 
    generated by BIM (one-to-one correspondence with the original example points). 
    If the generated adversarial sample is correctly predicted by the classifier, 
    the adversarial example is represented by $\bigstar$ (the corresponding original 
    example point is represented by $\bullet$); otherwise, the adversarial example is 
    represented by $\times$ (the corresponding original example point is represented 
    by $\square$). Note that Fig. 1 only shows the correctly predicted original example 
    points and their corresponding adversarial example points. It can be observed 
    that the iterative attack process of iterative adversarial attack algorithms 
    (BIM, PGD, etc.) gradually migrates (each subfigure in Fig. 1) the original 
    examples from the region $S_y$ to $S_k$ ($k\neq y$).

\subsection{Apply ACBI to adversarial robustness evaluation algorithm}

    \noindent As introduced in the previous introduction, the entire adversarial 
    attack process is to add a limited increment $\Delta v$ to the representation 
    vector $v$ so that $v+\Delta v$ gradually approaches the classification decision 
    boundary until it crosses the boundary (the attack is successful). 
    It can be seen from Fig. 1 that the example points closer to the 
    classification decision boundary are more vulnerable to attack. 
    Therefore, this paper proposes the ABCI method. Different from 
    the original random initialization (RI), ACBI uses the geometric 
    information of the representation space to purposefully initialize 
    the starting point of the attack and improves the effectiveness of 
    the attack algorithm. The difference between ACBI and RI is shown 
    in Fig. 2.
    \begin{figure}[!ht]
        \centering
        \includegraphics[width=0.5\textwidth]{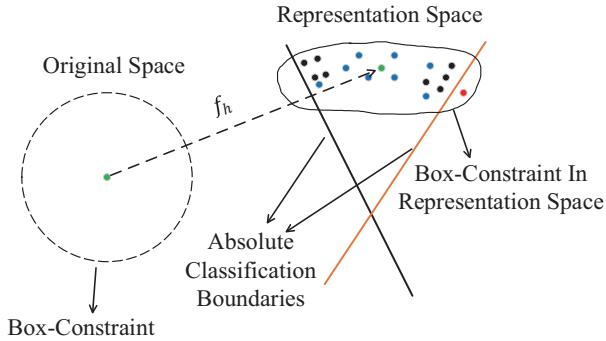}
        \vspace{-1em}
        \caption{Comparison of ACBI and RI}
        \label{Fig02}
    \end{figure}

    In Fig. 2, the left side shows the original input space, and the 
    right side shows the representation space of the original space 
    example points mapped by $f_h$. The blue points in the representation 
    space are generated by random initialization, the black points are 
    generated using the ACBI method, and the two lines represent the 
    absolute classification decision boundaries of the model. It can 
    be seen that the initial points generated by ACBI are better than 
    those generated by the RI method. They are not distributed around 
    the representation vector but near the absolute classification 
    boundaries, which will make it easier for the adversarial attack 
    algorithm to use these starting points to attack faster and more 
    efficiently.

    It is theoretically possible to obtain an adversarial example (
    represented as a red dot in Fig. 2) that is wrongly predicted 
    by the classifier through the absolute classification boundary, 
    but this is very difficult in practice. The mapping from the 
    original space to the representation space only needs to use 
    $f_h$. However, due to the existence of irreversible mapping 
    functions such as ReLu and MaxPooling, the reverse mapping 
    $f_h^{-1}$ from the representation space to the original space 
    cannot be described. Therefore, this paper uses ACBI to perform 
    multiple iterations to obtain an initial point closer to the 
    absolute classification boundary and then uses the original 
    attack algorithm for iterative attacks.

    First, this paper defines the distance from the representation 
    vector $v$ (assuming its true label belongs to the $y$-th class) to 
    the absolute classification boundary associated with the $k$-th 
    class ($k\in[0,K),k\neq y$) as:
        \begin{equation}
            \mathbb{D}(v,y,k)=\frac{F_{(y,k)}(v)}{\sqrt{\sum w_{(y,k)}^2}}
        \end{equation}
    
    The difference between Eq. (14) and the traditional geometric 
    distance definition is that it has a direction (i.e., $\mathbb{D}$ 
    may be a negative value), and the positive or negative of 
    $\mathbb{D}(v,y,k)$ is positive or negative depending on which side 
    of the line $F_{(y,k)}=0$ the representation vector $v$ falls on. 
    If $\mathbb{D}(v,y,k)<0$, it is indicated that $v$ falls into the $S_y$ 
    area, and the smaller the value, the farther it is from the 
    area $S_y$ where the label is located.

    In view of the above discussion, this paper gives the calculation 
    process of ACBI. In each restart, a starting point is randomly 
    selected to ensure the diversity of each restart:
        \begin{equation}
            x_{st}=x+\delta_0
        \end{equation}
    where $\delta_0\in U{(-\epsilon,+\epsilon)}^D$. Then, $x_{st}$ is 
    used as the iteration starting point to initialize using Eq. (14) 
    and the gradient descent method to obtain a start point that 
    closer to the classification decision boundary:
        \begin{equation}
            x_{init}^{t+1}={P_{(x,\epsilon)}(x}_{init}^t{-\eta}_{init}sign(\mathrm{\nabla}_{x_{init}^t}\mathbb{D}(v_{init}^t,y,m)))
        \end{equation}
    where,
        \begin{gather*}
        x_{init}^0=x_{st}\\
        v_{init}^t=f_h(x_{init}^t)\\
        m=\mathop{\mathrm{argmin}}\limits_{n\neq y}{\mathbb{D}(v_{init}^t,y,n)}
        \end{gather*}
    
    In Eq. (16), $\eta_{init}$ is the step size in the iterative process, 
    and function $P_{(x,\epsilon)}(\bullet)$ clips the input (the starting 
    point obtained by the $t$-th iteration) to the $\epsilon$-ball of $x$. 
    The representation vector will quickly reach its nearest classification 
    decision boundary through multiple initialization iterations.

    To demonstrate the effectiveness of ACBI, this paper uses PGD-4-25 
    (4 restarts, 25 iterations for each restart) with the ACBI method 
    for adversarial robustness evaluation on 10 defense models (including 
    5 network architectures, namely ResNet-18, WRN-28-10, WRN-34-10, 
    WRN-34-20, WRN-70-16, and 7 defense methods, namely DNR, OAAT, 
    Fixing Data, RLPE, FAT, ULAT, LBGAT). The results are shown in 
    Fig. 3.
        \begin{figure*}[!ht]
            \centering
            \includegraphics[width=1.0\textwidth]{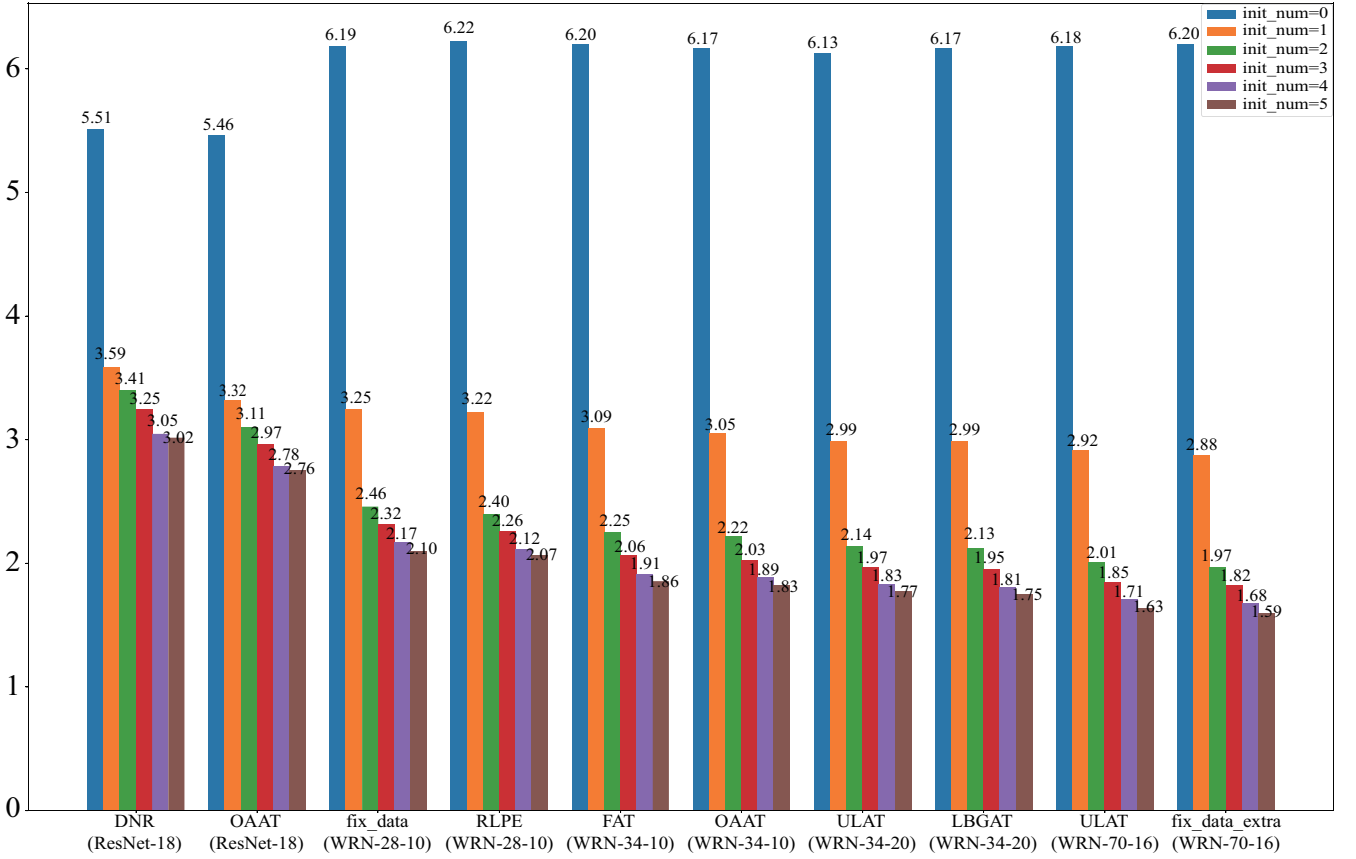}
            \vspace{-1em}
            \caption{Iteration number until attack success}
            \label{Fig03}
        \end{figure*}
    
    The experiments in Fig. (3) use ACBI with 0 to 5 initialization 
    iterations (when the number of ACBI initialization is 0, it is 
    the default version of PGD) on each defense model for attack 
    evaluation. Each bar counts the number of attacks required for 
    ACBI-PGD attacks to succeed under different initialization 
    iterations (mean of iteration numbers for a successful attack 
    in 1000 randomly selected examples from the CIFAR-10 test set). 
    It can be found that the more times the ACBI is initialized, the 
    fewer iterations are required for the attack to succeed (i.e., the 
    better attack starting point obtained by ACBI). That is, when 
    the number of attack iterations is fixed, the performance of the 
    attack algorithm using the ACBI method will be better (approaching 
    the lower bound of robustness).
    \begin{algorithm*}
        \caption{Improve Adversarial Attack Using ACBI.}
        \KwIn{
            Example pair$(x,y)$, 
            norm bound $\epsilon$, 
            the number of restarts $R$, 
            the number of iterations for ACBI $N_{init}$, 
            the step size for initialization iteration $\eta_{init}$ , 
            the number of the original attack iterations in each restart $N_{attack}$.
        }
        \KwOut{
            Adversarial example $x_{adv}^{t+1}$.
        }  
        \BlankLine
    
        \For{$r=0$ to $R$}{
            Using Eq.(15) to obtain the random start point $x_{st}^0$
            
            \For{$t=0$ to $N_{init}$}{
                Using Eq.(16) to obtain $x_{init}^{t+1}$
            }
            \For{$t=0$ to $N_{attack}$}{
                Using original attack algorithm in each iteration to obtain $x_{adv}^{t+1}$
            }
        }
    \end{algorithm*}

    Algorithm 1 shows how to apply ACBI to existing iterative attack algorithms. 
    Obviously, ACBI does not depend on existing attack algorithms and is 
    parameter-free (do not need to carefully adjust parameters). The specific 
    performance of ACBI will be shown in the experimental section.

\section{Experiment}

    \noindent This paper conducts comprehensive experiments on two datasets, 
    CIFAR-10 and CIFAR-100, to verify the effectiveness of the proposed method. 
    Two adversarial robustness evaluation methods, PGD and FAB, are used for 
    ACBI testing. The main reason is that the loss functions and iterative 
    process involved in the two attack algorithms are quite different, i.e., 
    these are two completely different types of attack algorithms. There are 8 
    different network structures, and nearly 50 $L-\infty$ defense algorithms are 
    used for adversarial robustness evaluation. Most of the algorithms are selected 
    from recent top conferences, e.g., ICML, NeurIPS, ICLR, ICCV, and CVPR. 
    Experiments show that the adversarial examples generated by the attack 
    algorithm using the ACBI method can make the classifier obtain lower 
    accuracy, indicating that using ACBI can improve the efficiency of the 
    adversarial robustness evaluation method.

    This paper uses the original PGD and FAB without the ACBI method as the 
    baseline. The restart number of all attack algorithms is set to 4, and the 
    number of attack iterations is set to 25. For a fair comparison, this paper 
    regards the number of initialization iterations as part of the number of 
    attack iterations, i.e., $N_{init}+N_{attack}=25$. The hyperparameter $\eta_{init}$ 
    is set to $\epsilon$, and there is no other careful tuning. The experimental 
    results are presented in Table 1 (CIFAR-10) and Table 2 (CIFAR-100).
\doublerulesep 0.1pt

\begin{table*}
\centering
\begin{footnotesize}
\caption{The performance of ACBI on CIFAR-10}
\begin{tabular}{
    p{0.1\textwidth}<{\centering} 
    p{0.2\textwidth}<{\centering} 
    p{0.05\textwidth}<{\centering} 
    p{0.05\textwidth}<{\centering} 
    p{0.15\textwidth}<{\centering} 
    p{0.05\textwidth}<{\centering}
    p{0.15\textwidth}<{\centering}
    }
\hline\hline

\multicolumn{2}{c}{CIFAR-10 Defense Model}&Clean&PGD&PGD$^*$&FAB&FAB$^*$\\
\hline

\multirow{7}{*}{ResNet-18}&Adv\_regular\cite{21}&90.58\%&74.09\%&61.48\%(-12.61\%)&21.61\%&17.53\%(-4.08\%)\\
&FBTF\cite{22}&87.85\%&60.93\%&55.25\%(-5.68\%)&34.15\%&32.34\%(-1.81\%)\\
&CNL\cite{23}\ddag&85.45\%&66.90\%&51.01\%(-15.89\%)&36.98\%&35.43\%(-1.55\%)\\
&Understanding FAST\cite{24}&84.38\%&62.57\%&51.00\%(-11.57\%)&39.10\%&38.11\%(-0.99\%)\\
&Proxy\cite{25}&84.54\%&62.02\%&53.03\%(-8.99\%)&42.54\%&41.87\%(-0.67\%)\\
&DNR\cite{26}&85.04\%&59.37\%&52.16\%(-7.21\%)&42.72\%&42.27\%(-0.45\%)\\
&OAAT\cite{27}&84.34\%&58.99\%&52.93\%(-6.06\%)&43.95\%&43.62\%(-0.33\%)\\
\hline

WRN-28-4&mma\cite{28}&86.75\%&61.07\%&60.29\%(-0.78\%)&57.60\%&57.52\%(-0.08\%)\\
\hline

ResNet-50&robustness\cite{29}&86.78\%&60.84\%&60.11\%(-0.73\%)&57.57\%&57.50\%(-0.07\%)\\
\hline

\multirow{11}{*}{WRN-28-10}&MART\cite{30}\dag&84.67\%&59.54\%&54.34\%(-5.2\%)&45.82\%&45.55\%(-0.27\%)\\
&Feature\_Scatter\cite{31}&85.26\%&60.79\%&55.93\%(-4.86\%)&48.03\%&47.76\%(-0.27\%)\\
&Adv\_inter\cite{32}&85.74\%&62.01\%&57.57\%(-4.44\%)&50.22\%&49.92\%(-0.3\%)\\
&awp\cite{33}\dag&85.90\%&62.29\%&58.30\%(-3.99\%)&51.22\%&50.95\%(-0.27\%)\\
&geometry\cite{17}\dag\ddag&86.20\%&62.75\%&59.17\%(-3.58\%)&51.95\%&51.70\%(-0.25\%)\\
&hydra\cite{34}\dag&86.40\%&62.66\%&59.42\%(-3.24\%)&52.87\%&52.65\%(-0.22\%)\\
&Pre-train\cite{35}&86.41\%&62.45\%&59.53\%(-2.92\%)&53.01\%&52.82\%(-0.19\%)\\
&rst\cite{36}\dag&86.58\%&62.56\%&59.89\%(-2.67\%)&53.46\%&53.29\%(-0.17\%)\\
&ULAT\cite{37}\dag&86.76\%&62.83\%&60.37\%(-2.46\%)&54.18\%&54.02\%(-0.16\%)\\
&Fix\_data\cite{38}&86.84\%&62.93\%&60.64\%(-2.29\%)&54.56\%&54.40\%(-0.16\%)\\
&RLPE\cite{39}\dag&86.98\%&62.94\%&60.82\%(-2.12\%)&54.92\%&54.76\%(-0.16\%)\\
\hline

\multirow{10}{*}{WRN-34-10}&TRADES\cite{19}\ddag&86.87\%&62.59\%&60.61\%(-1.98\%)&54.83\%&54.67\%(-0.16\%)\\
&awp\cite{33}&86.82\%&62.43\%&60.57\%(-1.86\%)&55.10\%&54.98\%(-0.12\%)\\
&Proxy\_dist\cite{25}\dag&86.80\%&62.50\%&60.76\%(-1.74\%)&55.25\%&55.19\%(-0.06\%)\\
&Self\_adaptive\cite{40}\ddag&86.65\%&62.29\%&60.64\%(-1.65\%)&55.19\%&55.12\%(-0.07\%)\\
&sensible\cite{41}&86.81\%&62.24\%&60.70\%(-1.54\%)&55.14\%&55.01\%(-0.13\%)\\
&yopo\cite{42}&86.85\%&61.58\%&60.15\%(-1.43\%)&54.78\%&54.70\%(-0.08\%)\\
&IAR/SAT\cite{43}&86.88\%&61.30\%&59.95\%(-1.35\%)&54.64\%&54.60\%(-0.04\%)\\
&LBGAT\cite{44}\ddag&86.94\%&61.05\%&59.78\%(-1.27\%)&54.71\%&54.57\%(-0.14\%)\\
&FAT\cite{45}&86.85\%&60.91\%&59.72\%(-1.19\%)&54.76\%&54.64\%(-0.12\%)\\
&OAAT\cite{27}&86.76\%&61.03\%&59.90\%(-1.13\%)&54.98\%&54.87\%(-0.11\%)\\
\hline

WRN-34-15&RLPE\cite{39}\dag&86.76\%&60.94\%&60.23\%(-0.71\%)&57.68\%&57.61\%(-0.07\%)\\
\hline

\multirow{4}{*}{WRN-34-20}&Hyper-embe\cite{46}&86.68\%&61.07\%&59.97\%(-1.1\%)&55.04\%&54.94\%(-0.10\%)\\
&Overfit\cite{47}&86.64\%&60.99\%&59.97\%(-1.02\%)&55.03\%&54.94\%(-0.09\%)\\
&ULAT\cite{37}&86.59\%&61.00\%&60.02\%(-0.98\%)&55.14\%&55.05\%(-0.09\%)\\
&LBGAT\cite{44}\ddag&86.63\%&60.88\%&59.96\%(-0.92\%)&55.17\%&55.05\%(-0.12\%)\\
\hline

\multirow{3}{*}{WRN-70-16}&ULAT\cite{37}&86.74\%&61.12\%&60.30\%(-0.82\%)&56.94\%&56.85\%(-0.09\%)\\
&ULAT\cite{37}\dag&86.78\%&61.15\%&60.28\%(-0.87\%)&56.50\%&56.40\%(-0.10\%)\\
&Fix\_data\cite{38}&86.79\%&61.30\%&60.52\%(-0.78\%)&57.70\%&57.62\%(-0.08\%)\\

\hline\hline

\end{tabular}
\end{footnotesize}
\end{table*}

    Table 1 shows the performance of ACBI on PGD-4-25 and FAB-4-25. The first column 
    shows the network architecture and the method name of the defense model (the one 
    marked with \ddag \ is $\epsilon=0.031$, and the one marked with \dag \ is an additional unlabeled 
    dataset used for training); the second column shows the original test example 
    accuracy (Clean). PGD* denotes PGD using ACBI strategy with $N_{init}=5$ and 
    $N_{attack}=20$; FAB* denotes the FAB using ACBI strategy with $N_{init}=1$ and 
    $N_{attack}=24$. $\epsilon$ in the attack process is set to 8.0/255, and the $\eta_{init}$ 
    in ACBI is equal to $\epsilon$. In the experiment, 1000 examples were randomly 
    selected for testing on CIFAR-10. The experimental results show that ACBI 
    obtains different degrees of improvement on PGD and FAB. The smaller the 
    number of model parameters, the better the ACBI performance (e.g., 
    ResNet-18 and WRN-28-10). One of the main reasons for this phenomenon 
    is that the larger the number of parameters (i.e., the more complex 
    $f_h$, the stronger the non-linearity of $f_h$), the more difficult 
    it is to optimize the distance from the starting point to the absolute 
    classification decision boundary using Eq. (16). This is reflected 
    in Table 2.

    In addition, the improvement brought by ACBI to FAB is significantly smaller 
    than that of PGD, which is related to the attack principle of FAB. FAB regards 
    the classifier as an approximately linear function. It uses the first-order Taylor 
    expansion at the example point x (true label is y) to approximate the decision 
    plane $\pi_s$ closest to $x$. Then, it uses ${Proj}_p(x_{adv}^{(i)},\pi_s,y)$ operation 
    to project the adversarial example $x_{adv}^{(i)}$ of the $i$-th iteration onto the 
    hyperplane and then clips to $[0,1]^D$. Therefore, the classification decision boundary 
    of the model is used in the FAB attack. Although it is approximately estimated 
    (fuzzy), it can still reflect the classification boundary information to a certain 
    extent, so the improvement of ACBI on FAB is lower than that of PGD.

    In this paper, the adversarial robustness evaluation performance of ACBI to 
    PGD and FAB on CIFAR-100 is presented in Table 2. The settings of various 
    parameters are consistent with CIFAR-10. The experimental results show 
    that ACBI obtains different degrees of improvement for different types 
    of attack algorithms on more complex data sets.
\doublerulesep 0.1pt
\begin{table*}
\centering
\begin{footnotesize}
\caption{The performance of ACBI on CIFAR-100}
\begin{tabular}{
    p{0.1\textwidth}<{\centering} 
    p{0.2\textwidth}<{\centering} 
    p{0.05\textwidth}<{\centering} 
    p{0.05\textwidth}<{\centering} 
    p{0.15\textwidth}<{\centering} 
    p{0.05\textwidth}<{\centering}
    p{0.15\textwidth}<{\centering}
    }
\hline\hline

\multicolumn{2}{c}{CIFAR-100 Defense Model}&Clean&PGD&PGD$^*$&FAB&FAB$^*$\\
\hline

\multirow{2}{*}{ResNet-18}&overfit\cite{47}&51.76\%&18.39\%&17.71\%(-0.68\%)&21.87\%&19.53\%(-2.34\%)\\
&OAAT\cite{27}&57.09\%&25.20\%&22.53\%(-2.67\%)&23.43\%&22.65\%(-0.78\%)\\
\hline

\multirow{2}{*}{WRN-28-10}&Pre-train\cite{35}&58.36\%&28.35\%&25.50\%(-2.85\%)&26.04\%&25.26\%(-0.78\%)\\
&Fix-data\cite{38}&59.62\%&30.24\%&27.58\%(-2.66\%)&27.14\%&26.95\%(-0.19\%)\\
\hline

\multirow{4}{*}{WRN-34-10}&awp\cite{33}&59.74\%&30.66\%&27.77\%(-2.89\%)&26.25\%&26.09\%(-0.16\%)\\
&IAR/SAT\cite{43}&59.93\%&30.01\%&27.36\%(-2.65\%)&25.26\%&25.13\%(-0.13\%)\\
&LBGAT\cite{44}\ddag&59.78\%&30.75\%&28.01\%(-2.74\%)&26.56\%&26.45\%(-0.11\%)\\
&OAAT\cite{27}&60.63\%&31.60\%&28.76\%(-2.84\%)&27.24\%&27.14\%(-0.10\%)\\
\hline

WRN-34-20&LBGAT\cite{44}\ddag&60.97\%&32.07\%&29.09\%(-2.98\%)&27.77\%&27.69\%(-0.08\%)\\
\hline

\multirow{3}{*}{WRN-70-16}&ULAT\cite{37}&60.72\%&32.71\%&29.66\%(-3.05\%)&28.36\%&28.30\%(-0.06\%)\\
&ULAT\cite{37}\dag&62.17\%&34.14\%&31.43\%(-2.71\%)&31.25\%&31.20\%(-0.05\%)\\
&Fix-data\cite{38}&62.05\%&34.31\%&31.81\%(-2.50\%)&30.95\%&30.91\%(-0.04\%)\\

\hline\hline
\end{tabular}
\end{footnotesize}
\end{table*}

    All experiments in this paper did not deliberately adjust the 
    parameters, and the experimental results are the average of five 
    groups of repeated experiments. ACBI can be integrated as a module 
    to other attack algorithms, e.g., BIM, CW, etc. Importantly, ACBI 
    and other strategies that improve the reliability of robustness 
    evaluation methods are highly loosely coupled, and they can be used 
    simultaneously in one attack algorithm.

\section{Conclusion}

    \noindent This paper first proposes the decouple space method to 
    divide the classifier $f$ into two parts, the non-linear mapping $f_h$ 
    and the linear mapping $f_t$. The original example can obtain its 
    representation vector $v$ on the $R^N$ space (referred to as the 
    representation space in this paper) through the mapping of $f_h$, 
    and then all linear problem-solving methods can be applied to $v$. 
    Then, from the perspective of decouple space, this paper uses $f$ to 
    define the absolute classification decision boundaries. Unlike the 
    fuzzy boundaries, given any classifier, this paper can obtain the 
    expressions of the absolute classification decision boundaries on 
    $R^N$. Next, ACBI is adopted to make the representation vector as 
    close as possible to the nearest classification decision boundary 
    through iterative optimization, thus obtaining a better attack 
    starting point. Finally, the experimental results on more than 50 
    defense models show that the adversarial robustness evaluation method 
    using ACBI is more reliable than the original ones.

\end{document}